\documentclass[%
 preprint,
 amsmath,amssymb,
  superscriptaddress,
]{revtex4-1}%

\usepackage{mhchem}
\usepackage[utf8]{inputenc} %
\usepackage[T1]{fontenc}    %
\usepackage{hyperref}       %
\usepackage{url}            %
\usepackage{booktabs}       %
\usepackage{amsfonts}       %
\usepackage{nicefrac}       %
\usepackage{microtype}      %
\usepackage{makecell}
\usepackage{amsmath}
\usepackage{amssymb}
\usepackage{color}
\usepackage{epstopdf}
\usepackage{physics}
\usepackage{enumitem}
\usepackage{graphicx}
\usepackage{subcaption}
\usepackage[yyyymmdd, 24hr]{datetime}

\begin{document}

\author{Tomohiko Konno}
\email[Corresponding author: ]{tomohiko@nict.go.jp}
\affiliation{National Institute of Information and Communications Technology}
\author{ Hodaka Kurokawa}\email[Second Corresponding author: ]{scottie0018@gmail.com}
\affiliation{The University of Tokyo}
\author{Fuyuki Nabeshima}
\affiliation{The University of Tokyo}
\author{Yuki Sakishita}
\affiliation{The University of Tokyo}
\author{Ryo Ogawa}
\affiliation{The University of Tokyo}
\author{Iwao Hosako}
\affiliation{National Institute of Information and Communications Technology}
\author{Atsutaka Maeda}
\affiliation{The University of Tokyo}

\title{Deep Learning Model for Finding New Superconductors} %

\begin{abstract}
Exploration of new superconductors still relies on the experience and intuition of experts, and is largely a process of experimental trial and error. In one study, only 3\% of the candidate materials showed superconductivity~\cite{1468-6996-16-3-033503}.
Here, we report the first deep learning model for finding new superconductors. We introduced the method named ``reading periodic table'' that represented the periodic table in a way that allows deep learning to learn to read the periodic table and to learn the law of elements for the purpose of discovering novel superconductors which are outside the training data. It is recognized that it is difficult for deep learning to predict something outside the training data. Although we used only the chemical composition of materials as information, we obtained an $R^{2}$ value of 0.92 for predicting $T_\text{c}$ for materials in a database of superconductors. We also introduced the method named ``garbage-in'' to create synthetic data of non-superconductors that do not exist. Non-superconductors are not reported, but the data must be required for deep learning to distinguish between superconductors and non-superconductors. We obtained three remarkable results. The deep learning can predict superconductivity for a material with a precision of 62\%, which shows the usefulness of the model; it found the recently discovered superconductor \ce{CaBi2} and another one \ce{Hf_{0.5}Nb_{0.2}V2Zr_{0.3}}, neither of which is in the superconductor database; and it found Fe-based high-temperature superconductors (discovered in 2008) from the training data before 2008. These results open the way for the discovery of new high-temperature superconductor families. The candidate materials list, data, and method are openly available from the \href{https://github.com/tomo835g/Deep-Learning-to-find-Superconductors}{link}.
\end{abstract}
\maketitle

\section{Introduction}

Extensive research has been conducted on superconductors with a high superconducting transition temperature, $T_\text{c}$, because of their many promising applications, such as low-loss power cables, powerful electromagnets, and fast digital circuits.
However, finding new superconductors is very difficult. In one study, it was reported~\cite{1468-6996-16-3-033503} that only 3\% of candidate materials showed superconductivity. 
Theoretical approaches have been proposed for predicting new superconducting materials. According to Bardeen-Cooper-Schrieffer (BCS) theory~\cite{PhysRev.106.162}, which explains phonon-mediated superconductivity in many materials, high $T_\text{c}$ is expected for compounds made of light elements. $T_\text{c}$ values of over 200 K have been reported for sulfur hydride~\cite{drozdov2015conventional} and lanthanum hydride~\cite{PhysRevLett.122.027001}. However, very high pressures (over 150 GPa) are required. Superconductivity with a rather high $T_\text{c}$ has been observed for cuprates~\cite{bednorz1986possible} and iron-based materials~\cite{kamihara2008iron} at ambient pressure, where unconventional superconductivity beyond the BCS framework is realized.  However, the strong electron correlations in these materials make it very difficult to conduct first-principles calculations~\cite{kohn1965self, jain2013commentary,kirklin2015open,curtarolo2012aflowlib} to calculate their electronic structures and predict their $T_\text{c}$ values. Therefore, new approaches for finding superconductors are needed. 
Materials informatics, which applies the principles of informatics to materials science, has attracted much interest~\cite{butler2018machine,ramprasad2017machine,HIMANEN2020106949,doi:10.1002/advs.201900808}. Among machine learning methods, deep learning has achieved great progress. Deep learning has been used to classify images~\cite{krizhevsky2012imagenet}, generate images~\cite{goodfellow2014generative}, play Go~\cite{silver2016mastering}, translate languages~\cite{Vaswani2017AttentionIA}, perform natural language tasks~\cite{Peters:2018}, and make its own network architecture~\cite{2018arXiv180203268P,Liu2018DARTSDA}. To predict the properties of materials using the conventional methods in materials informatics, researchers must design the input features of the materials; this is called feature engineering. It is very difficult for a human to design the appropriate features. A deep learning method can design and optimize features, giving it higher representation capabilities and potential compared to those of conventional methods. 
Many studies have been reported on drug discovery and organic chemistry by deep learning (mainly by graph neural networks~\cite{Zhou2018GraphNN,Wu2019ACS}), and on molecules~\cite{Elton2019DeepLF,Schtt2017SchNetAC}. %
Our results show the possibility of application of deep learning to inorganic materials and condensed matter physics, as additional areas outside organic chemistry.

\section{Reading periodic table}
\begin{figure}[htbp]
    \begin{subfigure}[t]{0.88\textwidth}
        \centering
        \includegraphics[width=1\textwidth]{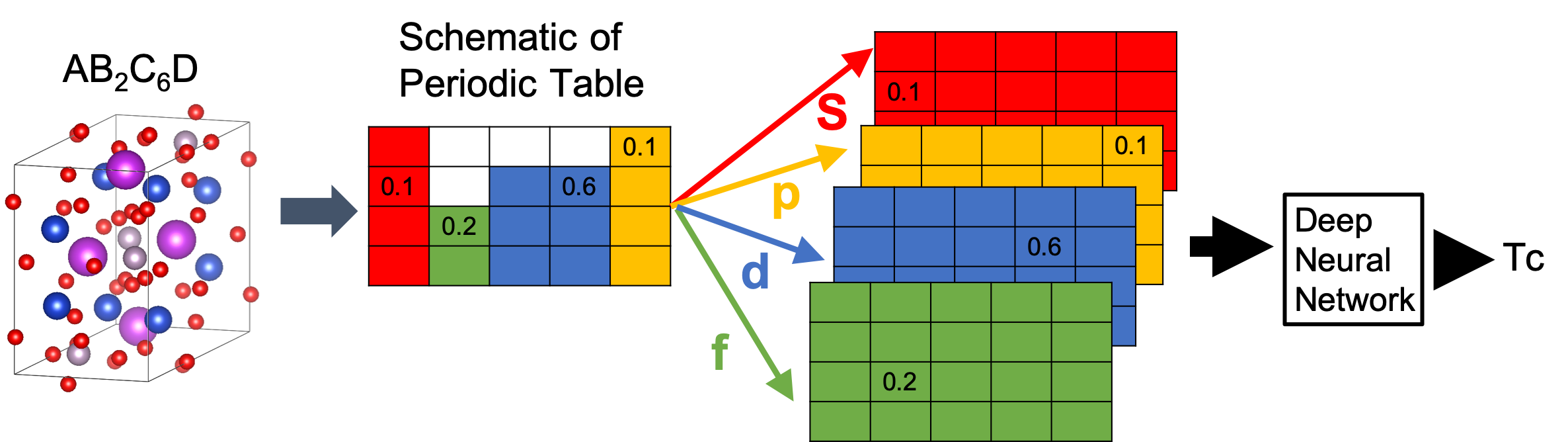}
    \end{subfigure} \\
    \vspace{2.5mm}
    \begin{subfigure}[t]{0.88\textwidth}
        \centering
        \includegraphics[width=1\textwidth]{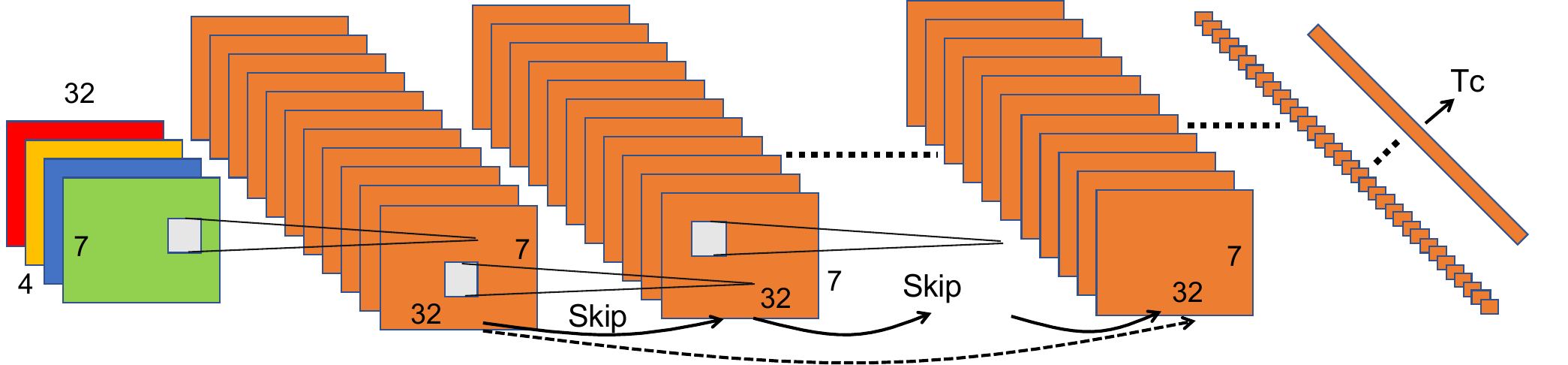}
	\end{subfigure}
	\caption{Proposed method named reading the period table. Top: the representation of a material by the method. The composition ratios of the material~\cite{vesta3} are entered into the two-dimensional periodic table. We then divide the original table into four tables corresponding to s-, p-, d-, and f-blocks, which show the orbital characteristics of the valence electrons, to allow the deep learning model to learn the valence orbital blocks. The dimensions of the representation are $4\times32 \times 7$. The neural network learns the rules from the periodic table by convolutional layers. Bottom: the representation by the method and neural network.}\label{fig: peridoc-table-neural-networks}
\end{figure}

Here, we report a deep learning model for the exploration of new superconductors.
Using deep learning to discover new superconductor families from known ones is analogous to using deep learning to recognize dogs from training data containing only cats.  This form of learning, called zero-shot learning, is very difficult. However, that the properties of elements can be learned by deep learning is shown by us, and they can be applied to materials. Our strategy is to suitably represent these properties, use this representation as training data, and have the deep learning model learn these properties. We made the deep learning model learn how to read the periodic table as human experts do. Although humans cannot recall tens of thousands of data points, computers can. For this purpose, we represented the periodic table in a way that allows a deep learning model to learn it, as illustrated in Fig.~\ref{fig: peridoc-table-neural-networks}. The convolutional layers learn the relative positions of the elements on the table, because they use the same local weights to whole periodic table. This is the reason why we use convolutional layers. Full connection layers should be basically avoided, since over-fitting easily occurs, and they do not learn the relative relationship. This method, named reading periodic table, is our first contribution to deep learning. We considered inorganic crystal superconductors because the number of known organic superconductors is small. We used only the composition of materials because the applied superconductor database does not have sufficient spatial information. (See more detail in Supplementary Information.)

\begin{figure}[htbp]
    \begin{center}
        \includegraphics[width=.7\hsize]{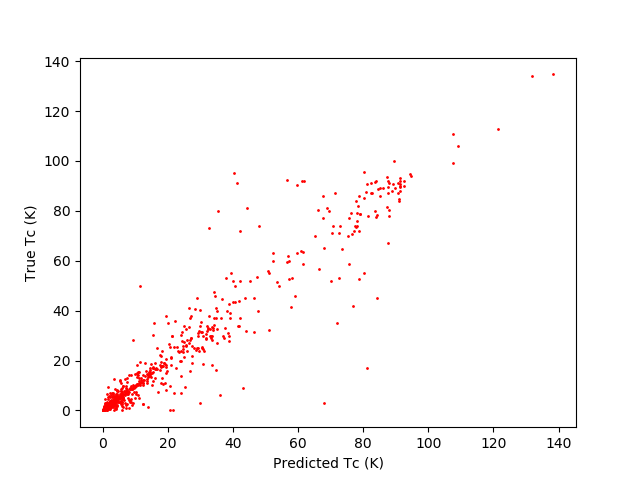}
        \caption{Scatter plot of predicted and true (SuperCon) $T_\text{c}$ values.}\label{fig: scatter_supercon}%
        \end{center}
    \end{figure}

We used the deep learning model to predict the critical temperatures, $T_\text{c}$, of superconductors in the SuperCon dataset~\cite{supercon}, which has the $T_\text{c}$ values of about 13,000 superconductors. We refer to the model trained with only SuperCon as the preliminary model. The train-test split was 0.05.  A scatter plot of the predicted and actual $T_\text{c}$ values is shown in Fig.~\ref{fig: scatter_supercon}. The $R^2$ value is 0.92, which is higher than that previously reported (0.88) for a random forest regression model~\cite{stanev2018machine}, where materials were restricted to those with $T_\text{c}> 10$ K (half of all materials). In contrast, our preliminary model does not have any restrictions regarding $T_\text{c}$ (see Supplementary Information). The random forest requires many appropriate input features of the materials (e.g., atomic mass, band gap, atomic configuration, melting temperature) to be manually designed. Here, even without such feature engineering, we achieved much better results.
\section{The problem in using data of superconductors only and the method named garbage-in for overcoming it}
We used the preliminary model trained with SuperCon to predict the $T_\text{c}$ values of 48,000 inorganic materials in the Crystallography Open Database (COD) to find new superconductors for experiments. However, for about 17,000 of the materials, the predicted $T_\text{c}$ was > 10 K, which is unreasonable. The failure to find new superconductors by this preliminary model seems to originate from the fact that the training data (SuperCon) included only 60 non-superconductors; the preliminary model was thus unable to learn non-superconductors. Data on non-superconductors are needed to differentiate superconductors from non-superconductors. However, no such dataset is available. Hence, we created synthetic data on non-superconductors, supposing that the $T_\text{c}$ values of the inorganic materials in COD that are not in SuperCon are 0 K under the assumption that most of these materials do not become superconductors with finite $T_\text{c}$. We used the synthetic data and SuperCon as the training data. We refer to this data generation method as garbage-in, which is our second contribution to deep learning.

As demonstrated by the above results for the preliminary model, scores of tests using only superconductor data, SuperCon, are not good for evaluating models. Usually, density functional theory is applied for evaluation in materials informatics; however, density functional theory cannot be used to evaluate models, because it is very difficult to calculate $T_\text{c}$ for strongly correlated systems. A database of non-superconductors is thus necessary.

\begin{table}[htbp]
	\begin{center}
\begin{tabular}{c|ccccc}
	\hline 
	&Accuracy&Precision  &Recall  &f1  \\ 
    \hline 
    Baseline (0 K)	 & & 32\%  &--  & -- \\ 
        \hline 
      Our DL model Reg (0 K)	&76\%& 62\%  & 67\%  & 63\%  \\  \hline

        Our DL model Cls (0 K)	&78\%& 72\%  & 50\%  & 59\%  \\  \hline
    Random Forest Cls (0 K) &73\%&71\% & 27\%&39\% \\
        \hline 
        	\hline 
Baseline (10 K)	& & 10\% &--  &--  \\ 
	\hline 
    Our DL model Reg (10 K)	&95\%& 75\%  & 76\% & 75\%  \\ 
    \hline 
 Our DL model Cls (10 K)	&95\%& 76\%  & 77\%  & 77\%  \\ 
    \hline 
        Random Forest Cls (10 K)&92\%& 88\%  &26\%  & 40\%  \\ \hline
\end{tabular}
\caption{Scores for predictions of superconductivity for materials reported by Ref.~\cite{1468-6996-16-3-033503}. Reg and Cls are abbreviations for regression and classification, respectively.} \label{tab: aprf_hosono} 
\end{center}
\end{table}

\section{The prediction of superconductivity}
We applied a list of materials reported by Ref.~\cite{1468-6996-16-3-033503} to evaluate the models. The list has about 400 materials found since 2010; importantly, it includes 330 non-superconductors. To temporally separate the materials on the list from the training data, we used only the data added to SuperCon or COD before 2010 as training data. The temporal separation test scheme is better than a random split of training and test data. The training and test data may end up being very similar after a random split. The temporal separation is the same situation when we use deep learning model to find new materials.

 We investigated outliers in $T_\text{c}$ predictions (see Fig.~\ref{fig: scatter_supercon}) and found that the under- and overestimated materials are cuprates, which have high $T_\text{c}$ that are sensitive to small changes in the ratio of elements. 
The surprise was that our deep learning model was sufficiently capable to find the mistake in the database. Some outliers are due to wrongly recorded $T_\text{c}$ values in SuperCon (database of superconductors). Mistakes in data are common. The $R^{2}$ is sensitive to such outliers. To compare the capability of a model with expert predictions, we evaluated whether the model could predict superconductivity for the given materials. Hence, we will use precision, recall, and f1 for evaluation.  %
Randomly selecting a material from the list with  $T_\text{c}$ > 0 K yields a precision of 32\%. This is considered the baseline because all the materials on the list were expected to be superconductors before the experiments. For the model predicting  $T_\text{c}$ with respect to whether it would be higher than 0 K, the results had a precision of 62\%, an accuracy of 76\%, a recall of 67\%, and an f1 score of 63\%. This precision is about two times higher than the baseline (32\%), which is about 10 sigma above it. The AUC was 0.78. %
Another interesting threshold is 10 K because only a limited number of superconductors have $T_\text{c} > 10$ K. The deep learning method predicted materials as being above this $T_\text{c}$ threshold with a precision of 75\%, which is about seven times higher than the baseline random precision (10\%). The accuracy, recall, and f1 score were 95\%, 76\%, and 75\%, respectively. The AUC was 0.94. In contrast, the preliminary model, trained with SuperCon only, predicted that all the materials would be superconductors, even though the training data were up to the year 2018 (i.e., not temporally separated). A previous study~\cite{stanev2018machine} used a random forest method. We also performed random forest binary classification with garbage-in and deep learning binary classification, which classify materials in terms of whether the $T_\text{c}$ is beyond 0 K or not. The AUC were 0.78 and 0.96, respectively. 
The results, summarized in Table~\ref{tab: aprf_hosono}, demonstrate that our deep learning model has good capability to predict superconductivity and clearly outperformed the previous method of random forest. (See Supplementary Information).%

\section{The discovery of two superconductors \ce{CaBi2} and \ce{Hf_{0.5}Nb_{0.2}V2Zr_{0.3}} }
Next, we used the model to predict the $T_\text{c}$ values of the materials in COD. The number of materials predicted to be superconductors was different every time we trained the models from scratch, which is expected with deep learning. We made a search target list for the experiment. After we removed cuprates and Fe-based superconductors (FeSCs) from the list, we obtained 900 materials predicted to be superconductors with $T_\text{c} > 0$ K, 280 materials with $T_\text{c} > 4$ K, and 70 materials with $T_\text{c} > 10$ K, which is more reasonable compared to the results obtained using the preliminary model. These materials are candidates for new superconductors. Although the prediction results on materials reported by Hosono et al.\ show that the model is useful, experiments (currently under way) are required to validate the method. The list included \ce{CaBi2}, which was recently found to be a superconductor~\cite{C6CP02856J} and another superconductor, \ce{Hf_{0.5}Nb_{0.2}V2Zr_{0.3}}~\cite{material-69}. The two superconductors are not listed in SuperCon. We had not known these were superconductors beforehand. It can be concluded that the deep learning model found actual superconductors. We have made the list openly available.

Another interesting prediction regards \ce{BeB2}. The material \ce{MgB2} is a famous superconductor with $T_\text{c}=40$ K, and the element \ce{Be} is just above the element \ce{Mg} in the periodic table. In the used database, the number of two-element materials that include \ce{B} is more than $300$. Although \ce{BeB2} is not a superconductor, it is not a coincidence that deep learning predicted \ce{BeB2} as such. This is evidence that the deep learning model reads the periodic table to predict superconductors in a similar way as human do.

\section{The discovery of Fe-based superconductors (FeSCs)}
To test the capability of our deep learning model of finding new types of superconductor, we investigated whether we could find high-$T_\text{c}$ FeSCs by using the model trained with data before 2008, the year FeSCs were discovered. We removed two materials, \ce{LaFePO} and \ce{LaFePFO}, from the training data because their discovery in 2006 led to the discovery of high-$T_\text{c}$ FeSCs. We used the 1,399 FeSCs known as of 2018 in SuperCon as the test data. A total of about 130 training and test runs were used. Although the models were made stochastically, we found some FeSCs that were predicted to have finite $T_\text{c}$. A histogram of the number of predicted FeSCs with $T_\text{c}$ $> 0$ K is shown in Fig.~\ref{fig: hist_fe_2007_without-log-normal}. We obtained the same results for high-$T_\text{c}$ cuprates (see Supplementary Information). 
When we used shallow 10-layer networks that had as good $R^2$, precision, etc., as the current large model, FeSCs were not found. This is not strange, because most iron compounds show magnetism, which is incompatible with superconductivity, and there are few superconductors including iron except for FeSCs. Indeed, few researchers had anticipated that FeSCs could have high Tc values. It is recognized that larger models have better generalization performances. The fact that the larger model found FeSCs can be explained by a larger model having an improved search capability for new superconductors. 
We must mention that random forest models could not find FeSCs. 
These results suggest that FeSCs and cuprate superconductors might have been found by our deep learning model.

\begin{figure}[htpb]
       \centering
       \includegraphics[width=0.7\textwidth]{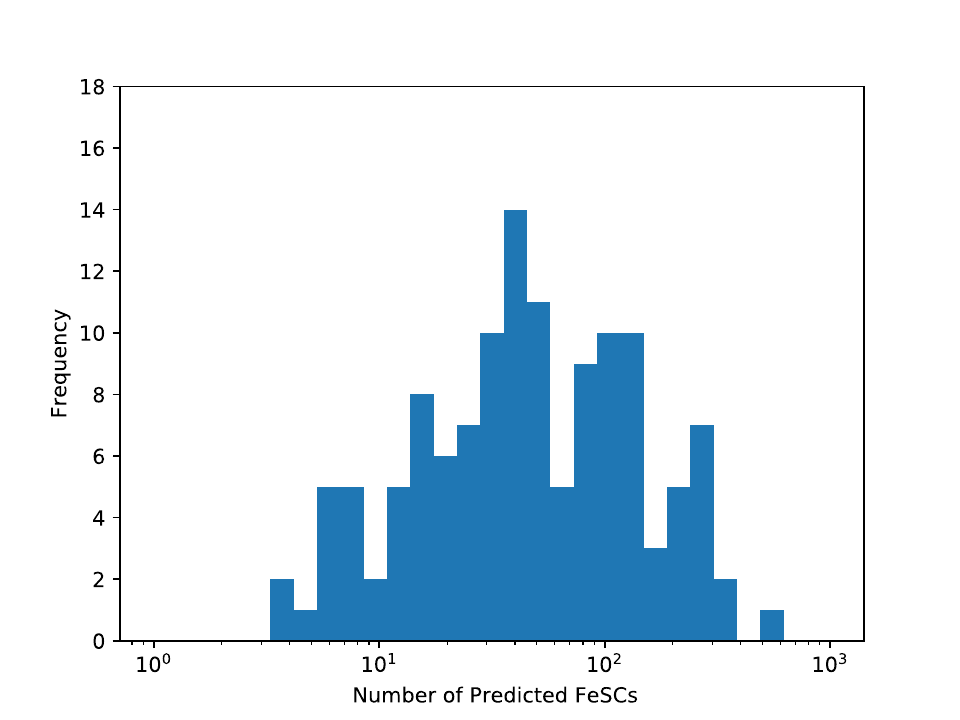}
	\caption{Histogram of the number of predicted FeSCs with $T_\text{c} > 0$ K (log scale).}\label{fig: hist_fe_2007_without-log-normal}
\end{figure}

The code and the data are available from the \href{https://github.com/tomo835g/Deep-Learning-to-find-Superconductors}{link}.

\section{Discussion}
If we had searched for FeSCs following the prediction, we would have discovered FeSCs. However, the predicted $T_\text{c}$ of the FeSCs was rather low in our attempt to \emph{discover} FeSCs. FeSCs might thus have been a low-priority target depending on how the model prediction was used. This problem will be considered in future research. We will incorporate crystal structure information to enhance the capability of the model of finding new high-$T_\text{c}$ superconductor families. Nevertheless, the present model is still useful as an auxiliary tool. Furthermore, the present method could be applied to other problems where crystal structure is difficult to obtain.

Even though our method does not require feature engineering, unlike conventional methods in materials informatics, it achieved much better results. Our deep learning method may replace existing methods, just as other deep learning methods have done in computer vision, natural language processing, and reinforcement learning. Deep learning requires failure data (e.g., non-superconductors) for accurate prediction. As many datasets in materials search are a random train-test split, we must prepare a temporally separated train-test datasets for the field to progress. Because our method does not use specific properties of superconductors and uses only chemical formulas, the method can be applied to other problems with, in particular, inorganic materials. We demonstrate band gap estimation by our method in Ref.~\cite{doi:10.7566/JPSJ.89.124006}. We demonstrated the usefulness of our method and deep learning to inorganic materials and condensed matter physics as areas outside organic chemistry, the studies of which have been much reported yet. Our results open the way to the discovery of new high-$T_\text{c}$ superconductor families, which must open up new physics. %

\section{Summary of introduced methods and the results}
The summary is given for readers.
\subsection{Summary of Introduced methods}
\begin{enumerate}%
\item The first deep learning model for finding new superconductors. 
\item Reading periodic table: the method that allows deep learning to learn to read the periodic table in order to learn the laws of elements. 
\item Garbage-in: the method to create synthetic data on non-superconductors. 
\item Model evaluation scheme that uses temporally separate training and test data.
\end{enumerate}

\subsection{Summary of Results}
\begin{enumerate}[start=0]
\item (Good $R^{2}$ value for estimating $T_\text{c}$ by using data of superconductors only.) 
\item The deep learning method predicted superconductivity for a material with a precision of 62\%. 
\item The deep learning method had better capabilities than random forest. 
\item The deep learning method discovered superconductors \ce{CaBi2} and  \ce{Hf_{0.5}Nb_{0.2}V2Zr_{0.3}}. 
\item The deep learning method found Fe-based high-temperature superconductors (discovered in 2008) from the training data before 2008. 
\end{enumerate}
\newpage

 \quad \\
\textit{Author Contributions}-- Tomohiko Konno conceived and supervised the research.
Tomohiko Konno, Hodaka Kurokawa, Yuki Sakishita, and Fuyuki Nabeshima had the primary roles. Tomohiko Konno, Hodaka Kurokawa, and Fuyuki Nabeshima discussed the direction and interpretation of the analysis, and were the main writers of the manuscript. 
Tomohiko Konno made the deep learning model and the two methods (reading periodic table and garbage-in), and specified how to evaluate a model using the materials reported by Hosono et al. and the temporal separation.
Hodaka Kurokawa checked the materials in the candidate materials list and found the two superconductors, and investigated the corresponding original papers to determine the indefinite values in the materials reported by Hosono et al. Yuki Sakishita performed random forest analysis. Iwao Hosako brought together the experimenters and machine learning experts. All authors approved the final version of the manuscript for submission.

\textit{Data availability}-- 
All the data used, SuperCon~\cite{supercon}, COD~\cite{gravzulis2011crystallography,gravzulis2009crystallography,Downs2003}, and the materials reported by Ref.~\cite{1468-6996-16-3-033503} are openly available. The materials reported by Hosono et al.\ have undetermined variables, such as $x$ in \ce{H_{2-x}O_{1+x}}. We investigated related papers and input the values for such variables. We then made a list of materials for the evaluation of models. This list will be openly available under the condition written in the following site for the community from the \href{https://github.com/tomo835g/Superconductors}{link}. See Supplementary Information also for data handling.

\textit{Correspondence}--
Correspondence and requests for materials should be addressed to Tomohiko Konno (tomohiko@nict.go.jp) and Hodaka Kurokawa (scottie0018@gmail.com).

\bibliography{bib_deep_matter.bib}

\newpage

{\Large Supplementary Information}
\appendix

\section{The method named reading periodic table: representation of periodic table}

\subsection{Representation of elements as one-hot vectors}\label{sec: one-hot-1}
Any one of the $118$ elements of the periodic table can be represented by a one-hot vector. For example, He can be represented by a $118$-dimensional vector $(0,1,0,\cdots, 0)$ and H can be represented by $(1,0,\cdots,0$). The fictional compound \ce{H2He3} would be represented by $(2,3,0,\cdots,0)$ or $(2/5,3/5,0,\cdots,0)$. %
There are two problems associated with representing materials by one-hot vectors. First, neural networks do not learn about elements and their combinations that do not appear in the training data. Second, one-hot vector representations do not reflect the properties of the elements, especially when data are scarce. Elements are treated as quite different entities in one-hot representations, even though the properties of the elements are known from quantum mechanics.

\subsection{Learning of periodic table}
To overcome these problems, we introduce a method that enables the deep learning model to learn the periodic table. The information on elements is reflected by the data representation, which the deep learning model uses to learn the properties. The properties of the elements and their similarities are reflected in the periodic table.%
The composition ratios of materials are entered into the periodic table and we then divide the periodic table into four tables corresponding to the s-, p-, d-, and f-blocks %
because differences in the valence orbitals are important. The dimensions of the representation are $4\times32 \times 7$. The deep learning model learns the periodic table using its convolutional layers. With knowledge of the periodic table and element properties, the deep learning model can predict unknown materials from known ones.

\begin{figure}%
    \begin{center}
        \includegraphics[width=0.8\hsize]{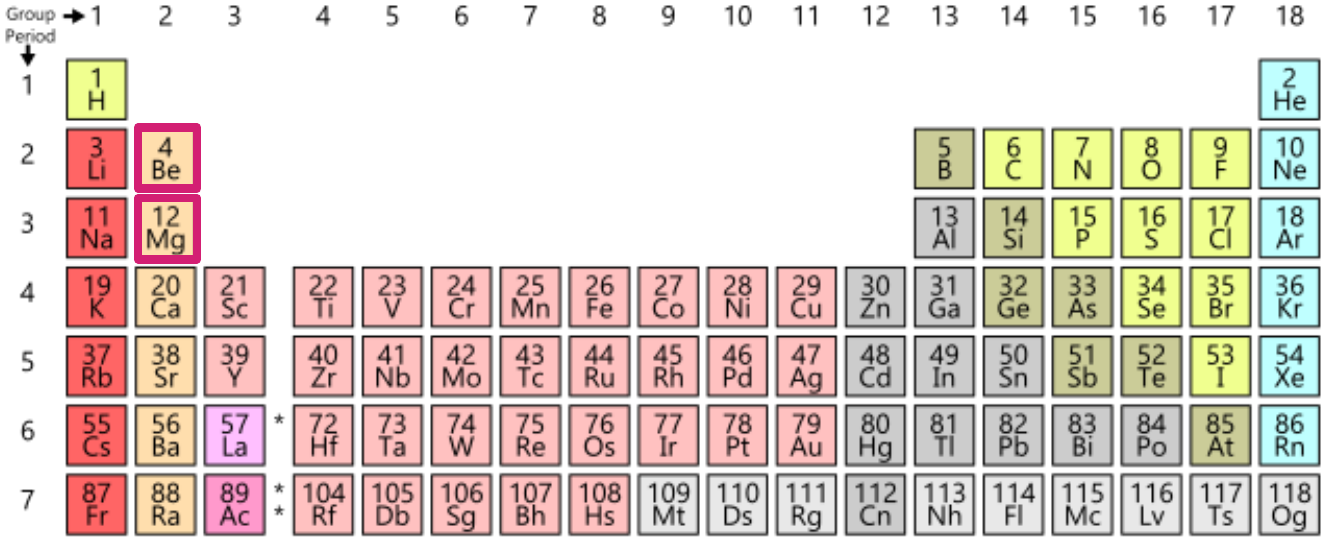}
        \caption{Be and Mg (left top) are highlighted by red frames.}\label{fig:BeMg}
        \end{center}
    \end{figure}

Here we show that our deep learning model and the reading periodic table method indeed read the periodic table and make predictions. As is described in the main manuscript, we had the deep learning model make a candidate superconductors list. In that list, we found \ce{BeB2}. The material \ce{MgB2} is a famous superconductor with $T_\text{c}=40$ K, and the element \ce{Be} is just above the element \ce{Mg} in the periodic table as shown in Fig.~\ref{fig:BeMg}. In the database used for prediction, the number of two-element materials that include \ce{B} is more than 300. Although \ce{BeB2} is not a superconductor, it is not a coincidence that deep learning predicted \ce{BeB2} as such. This shows that the deep learning model reads the periodic table to predict superconductors in a similar way as human do.

\section{Garbage-in: a method for creating synthetic data on non-superconductors} 
\begin{figure}%
    \begin{center}
        \includegraphics[width=0.8\hsize]{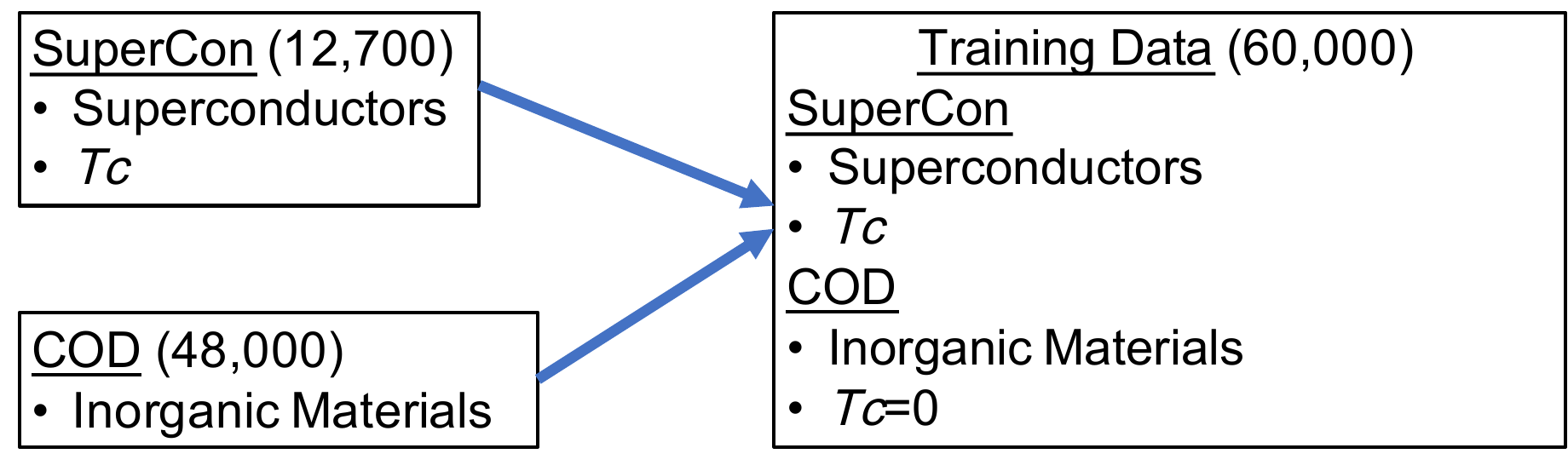}
        \caption{Synthetic data generation method named garbage-in.}\label{fig: garbage_in}
        \end{center}
    \end{figure}

\begin{figure}%
    \begin{center}
        \includegraphics[width=0.9\hsize]{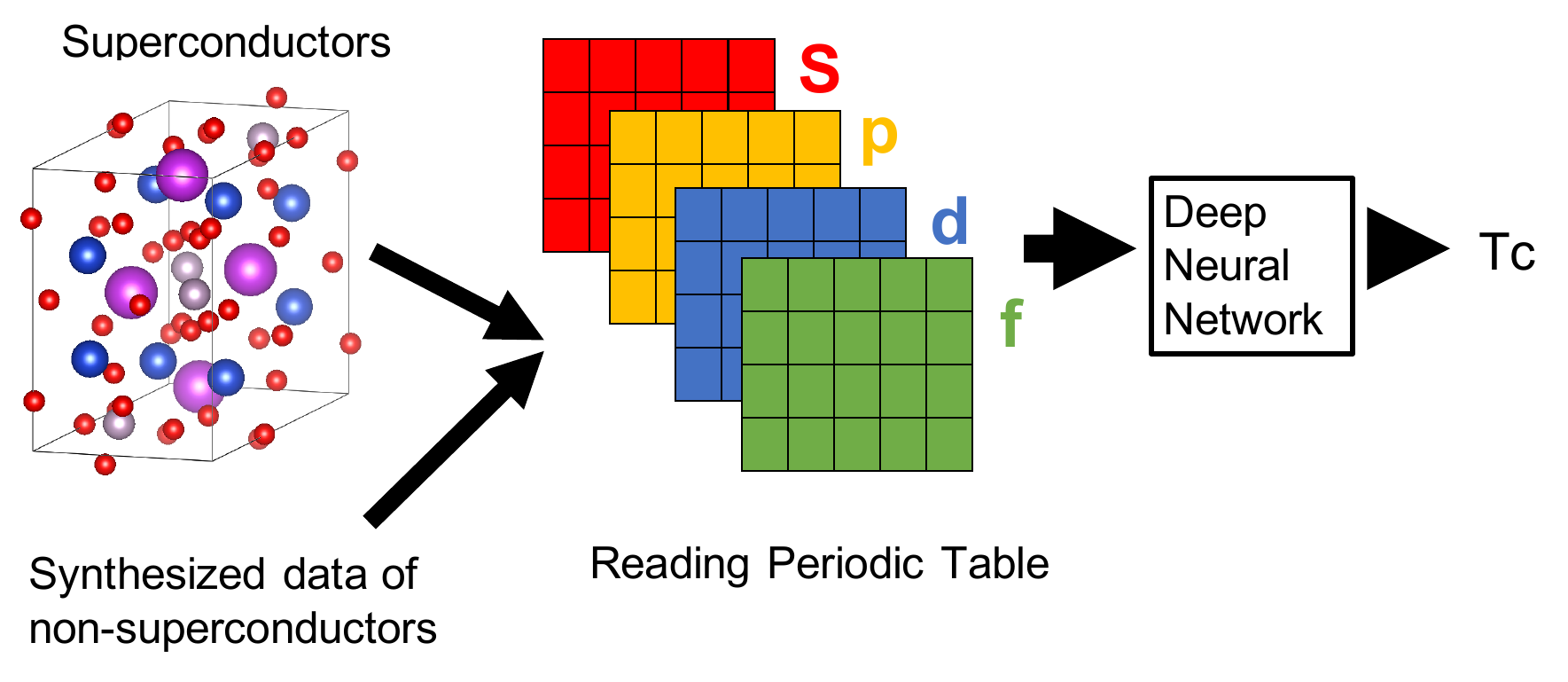}
        \caption{Overall scheme. The data of superconductors from SuperCon and the data of non-superconductors synthesized by garbage-in are transformed into the representation by reading periodic table in order for neural networks to learn the rules, then the deep neural network is trained to output $T_\text{c}$ }\label{fig: overall}
        \end{center}
    \end{figure}

We have a database of superconductors. However, to explore new superconductors, we also need a database of non-superconductors, which does not exist. Hence, we created synthetic data.
Under the assumption that most of the inorganic materials in COD do not become superconductors with finite $T_\text{c}$, we input the inorganic materials in COD with \(T_c=0\) to the deep learning model as training data along with SuperCon. The method is illustrated in Fig.~\ref{fig: garbage_in}. The overall scheme of training is illustrated in Fig.~\ref{fig: overall}.

\section{List of candidate materials}
We used the 48,000 inorganic materials in COD, 1,000 of which were used as test data. The remaining 47,000 materials and 12,000 materials in SuperCon were used as training data. Then, we obtained predicted $T_\text{c}$ values of the 1,000 materials in the test data. We repeated the procedure 48 times with different test data. This produced a candidate materials list. If we generate materials by generative models, which output chemical composition virtually, we do not yet know how to synthesize the generated materials. COD is thus used because it is a list of previously synthesized materials.

\section{Data availability} SuperCon~\cite{supercon}, COD~\cite{gravzulis2011crystallography,gravzulis2009crystallography,Downs2003} and the materials reported by Ref.~\cite{1468-6996-16-3-033503} are openly available and free to use. The materials reported by Hosono et al.\ have undetermined variables, such as $x$ in \ce{H_{2-x}O_{1+x}}. We investigated related papers and input the values for such variables. We then made a list of materials for the evaluation of models. This list will be openly available under the condition written in the following site for the community from the \href{https://github.com/tomo835g/Superconductors}{link}.
\section{Code and Candidate materials list}
The codes of methods are available from the \href{https://github.com/tomo835g/Deep-Learning-to-find-Superconductors}{link} under the condition written in the link. The candidate materials list is also available.

\section{Data handling}
\subsection{Definitions of conventional, cuprate, and Fe-based superconductors}
Cuprate superconductors are defined as materials that contain Cu, O, and more than one other element. The exceptions are \ce{Cu}, \ce{La}, and \ce{O}. FeSCs are defined as materials that contain Fe and either As, S, Se, or P. All other superconductors are considered to be conventional.
\subsection{Removal of problematic data}
We removed materials whose composition ends with variables such as ``$+$x''. About 7,000 materials were removed from SuperCon. If we know the accurate compositions of these materials and include them in the data, this should improve the deep learning model. We input appropriate values for variables such as ``x'' for the materials reported by Hosono et al.\ after reviewing the original studies, because there were only about 300 materials left after temporal separation.

\subsection{Treatment of materials with same composition but different $T_\text{c}$ values in SuperCon}
SuperCon contains materials with the same composition but different $T_\text{c}$ values. We decided to use the median value of $T_\text{c}$. %

\subsection{Treatment of materials without $T_\text{c}$ values in SuperCon}
Of the 17,000 remaining materials, about 4,000 did not have $T_\text{c}$ values. We considered setting $T_\text{c}=0$ for these materials or just excluding them. A comparison of the regression results of the preliminary models with SuperCon indicated that excluding the materials without $T_\text{c}$ values was better, so this was done. 
\subsection{Treatment of COD data}
We use only the inorganic materials in COD. We remove duplicates, data with compositions difficult for machines to read, and overlap with SuperCon and the materials reported by Hosono et al. After this process, about 48,000 materials remained. 
\subsection{Overlap among SuperCon, COD, and materials reported by Hosono et al.}
The overlap with SuperCon was removed from COD and the materials reported by Hosono et al. %
\section{Temporal separation of materials}
Since the materials reported by Hosono et al.\ were collected starting from 2010, we used data from before 2010 as the training data. Using data from before 2008 as the training data and using the materials reported by Hosono et al.\ to check the reliability of the models also resulted in temporal separation.

\section{Accuracy, precision, recall, and f1 score}
Consider disease detection. Suppose that we have 10,000 samples, 100 of which contain a disease. The task is to predict whether a given sample contains a disease. Accuracy is the rate of the prediction being right, irrespective of the prediction being positive or negative. Precision is the percentage of positive predictions that are correct. If a positive prediction is made for only one sample that is obviously positive and all other samples are predicted to be negative, then you will get 100\% precision but you will miss all the remaining $99$ samples with a disease, which is a problem for disease detection.
Hence, we have recall. Recall is the percentage of identified samples with a disease out of all samples with a disease. If all 10,000 samples are predicted to be positive, recall will be 100\% since all 100 samples with a disease were found, but accuracy and precision will be only 1\%, which is unsatisfactory. %
The f1 score utilizes both precision and recall. It is given by the harmonic mean $2\times\frac{\text{precision}\times\text{recall}}{\text{precision}+\text{recall}}$. The best measure for evaluating a model depends on the specific problem.

\section{Neural networks}
A smooth L1 loss function was used. The optimizer was Adam~\cite{Kingma2014AdamAM}. 
For the prediction of $T_\text{c}$ values for the materials in SuperCon by the preliminary model, the learning rate was $2\times 10^{-6}$, the batch size was 32, the number of epochs was 6,000, $T_\text{c}$ was in the linear scale, and the number of layers was 64.  %
For the prediction of $T_\text{c}$ values for the materials in SuperCon by the model with garbage-in, the number of epochs was set to 1,000. It took about 45 hours for training because the training dataset was five times larger than the preliminary model.
For the prediction of superconductivity for the materials reported by Hosono et al., the learning rate was $10^{-4}$, the batch size was 32, the number of epochs was 200, $T_\text{c}$ was in the linear scale, and the number of layers was 64. 
For the prediction of FeSCs, the learning rate was $10^{-4}$, the batch size was 32, the number of epochs was 200, $T_\text{c}$ was in the linear scale, and the number of layers was 64. For making the candidate materials list for the experiment and the discovery of \ce{CaBi2} and \ce{Hf_{0.5}Nb_{0.2}V2Zr_{0.3}} from the list, the learning rate was $10^{-4}$, the batch size was 32, the number of epochs was 500, $T_\text{c}$ was in the log scale after the addition of 0.1 to $T_\text{c}$, and the number of layers was 9. The network was different because these predictions were done at the start of our research. It took a significant amount of time to check the list and, consequently, find the material. We also found the superconductors \ce{CaBi2} and \ce{Hf_{0.5}Nb_{0.2}V2Zr_{0.3}} using the 64-layer network. %

\section{Random forest}
Random forest analysis was performed by using the weighted average, weighted variance, maximum, minimum, range, mode, median, and mean absolute difference of the 32 basic features of elements in compositions. The basic features were obtained from \href{https://bitbucket.org/wolverton/magpie/src/master/}{Magpie}. The results were averages over 10 models. Random forest analysis using only the data of superconductors, without garbage-in, encountered the same problem as our deep learning model. It predicted about 60\% of the materials were superconductors. We performed classification regarding whether $T_\text{c}$ is beyond 0 K or not for materials reported by Hosono et al. because it is almost impossible for random forest regression to estimate $T_\text{c}=0$, due to random forest classification being an ensemble estimation. If even one tree estimates $T_\text{c}>0$, then random forest regression estimates $T_\text{c}>0$.  Classification with respect to 10 K was also done.  For the random forest, we set the number of estimators as 100. 
\section{Summary of random forest capability}
The random forest models are dominated by our deep learning model in our test scheme for temporal separation and telling the difference between superconductors and non-superconductors. The random forest classification did not find any FeSCs. %

When good features are available, feature-based methods including random forest are better than deep learning in our experience. We suppose the following two points are possible reasons for the poor capability shown by random forest. First, the features conventionally used in materials and machine learning are not good features for superconductors. Superconductors, in particular for a strongly correlated system, leave much room to be investigated theoretically. Second, the conventional test scheme in materials and machine learning is a random train-test split scheme. Our temporal separation train-test data scheme revealed the dominance of deep learning.

\section{Training and test data used for main results}
The training data and test data used for the main results are summarized in Table~\ref{tab: summary-results}.
\begin{table}[htbp]
	\begin{center}
\begin{tabular}{|c|c|c|}
	\hline 
	Main result &Training data  &Test data  \\ \hline%
	\makecell{Prediction of superconductors from \\materials reported by Ref.~\cite{1468-6996-16-3-033503}}&\makecell{SuperCon and COD \\before 2010} &  Materials reported by Hosono et al.\\ \hline
	\makecell{Superconductors \ce{CaBi2} and \\  \ce{Hf_{0.5}Nb_{0.2}V2Zr_{0.3}}  found \\in candidate materials list}&SuperCon and COD in 2018& COD in 2018 \\ \hline
	\makecell{FeSCs found from training \\data before discovery year}&\makecell{SuperCon and COD \\before 2008}& \makecell{FeSCs in SuperCon in 2018} \\ \hline
\end{tabular}\caption{Summary of main results. } \label{tab: summary-results}
\end{center}
\end{table}

\quad \\
\newpage
\quad \\
\newpage
{\Large More Supplementary Information}
\section{Other hyper-parameters}
We obtained an $R^{2}$ value of 0.93 for the prediction of $T_\text{c}$ for the materials in SuperCon by the best preliminary model with the same train-test split (0.15) as that used in a previous study~\cite{stanev2018machine} by random forest, because in the previous study, only the $R^2$ value of the best model (to our understanding), 0.88, was reported, which is less than our $R^2$ value of 0.93. For a train-test split of 0.05, the median of $R^{2}$ was 0.92 for 56 models, which is presented in the main text. For the prediction of $T_\text{c}$ values for the materials in SuperCon by the model with garbage-in and a train-test split of 0.05, the median of $R^{2}$ was 0.85 for 55 models. For the prediction of superconductivity for the materials by deep learning regression reported by Ref.~\cite{1468-6996-16-3-033503}, the reported scores are the median values for 29 models. For deep learning classification, the scores are the averages over 32 models, and the hyper-parameters are the same as those of the regression models that output $T_\text{c}$ except for the binary classification and the binary classification entropy with the logit loss function. %

\section{Prediction of superconductivity in FeSCs by deep learning model}

The number of predicted FeSCs varied with the model because models trained with the same training data can become different depending on their initial weights and the input order of the training data. The models were stochastically constructed. However, once a model is constructed, the output is deterministic unless stochastic layers are used. %
Of note, we always found some high-$T_\text{c}$ FeSCs that were predicted to have a finite $T_\text{c}$. %

\subsection{Check of models based on materials reported by Hosono et al. }\label{sec: check-by-hosono}
A model that predicts superconductivity for all materials or makes random predictions will have a precision that is equal to the baseline random precision. The validity of the models was checked using the materials reported by Hosono et al., which can be used to reject the model.
We used the models to predict whether the materials on the list had a superconducting transition temperature of above 0 K, and checked whether the precision was higher than the baseline random precision. %
The mean precision was 0.5 for about 130 training and test runs and the baseline random precision was 0.32.  The precision was about two times higher than the baseline. We also checked whether each model satisfied the condition that the precision be sufficiently higher than the baseline. %

\subsection{Predictions using various combinations of training data}

 To confirm the reliability of a model, we checked whether the model learned the feature of superconductivity by checking the effect of training data on the number of predicted FeSCs. We compared the predictions of five models trained using different data based on SuperCon and COD. The training data were as follows: (i) data before 2008 without LaFePFO and LaFePO; (ii) data before 2008 with LaFePFO and LaFePO; (iii) only conventional superconductors as of 2018; (iv) only cuprates as of 2018; and (v) both conventional superconductors and cuprates as of 2018. These models predicted the $T_\text{c}$ values of the FeSCs in SuperCon. In total, about 130 training and test runs were used for (i) and (ii) and about 170 training and test runs were used for (iii), (iv), and (v). 

\begin{figure}[htpb]
    \begin{subfigure}[t]{0.5\textwidth}
        \centering
        \includegraphics[width=1.1\textwidth]{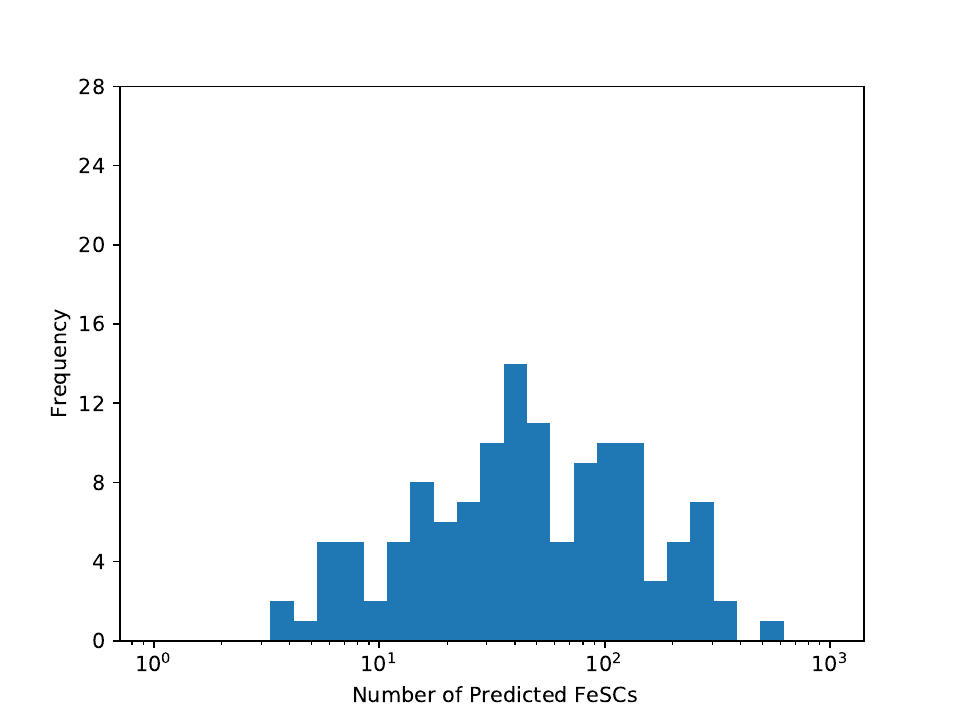}
            \caption{Without LaFePFO and LaFePO.}\label{fig: 2007-without}
    \end{subfigure}%
    \begin{subfigure}[t]{0.5\textwidth}
        \centering
        \includegraphics[width=1.1\textwidth]{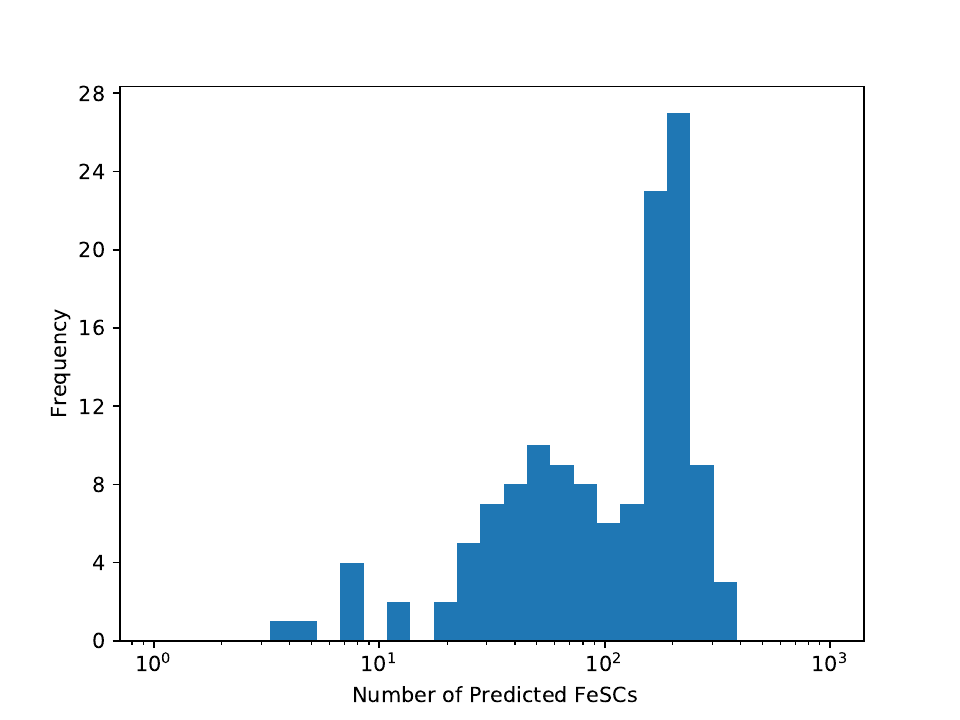}
            \caption{With LaFePFO and LaFePO.}\label{fig: 2007-with}
    \end{subfigure}
       \caption{Comparison of histograms of predicted FeSCs by model trained with data before 2008 with and without LaFePFO and LaFePO (log scale).}
    \label{fig: 2007-with-without}
    \end{figure}

Figure~\ref{fig: 2007-with-without} shows the results for (i) and (ii). As shown, the number of predicted FeSCs in Fig.~\ref{fig: 2007-with} is higher than that in Fig.~\ref{fig: 2007-without}. The average number of FeSCs predicted to have finite $T_\text{c}$ increased from 80 to 129. The median increased from 44 to 130. We checked the validity of the models using the materials reported by Hosono et al. When evaluating model (i), we removed materials containing Fe from the list. This was not done when evaluating model (ii). The baseline random precision and model precision were 0.32 and 0.5, respectively, for model (i), and 0.32 and 0.5, respectively, for model (ii). The model precision was sufficiently higher than the baseline, indicating that the models were valid. Because we did not use full data as the training data, the precisions were smaller than the value of 62 \% reported yet. It is also true for the rest other models (iii), (iv), and (v).

\begin{figure}[htpb]
    \begin{subfigure}[t]{0.5\textwidth}
        \centering
        \includegraphics[width=1.1\textwidth]{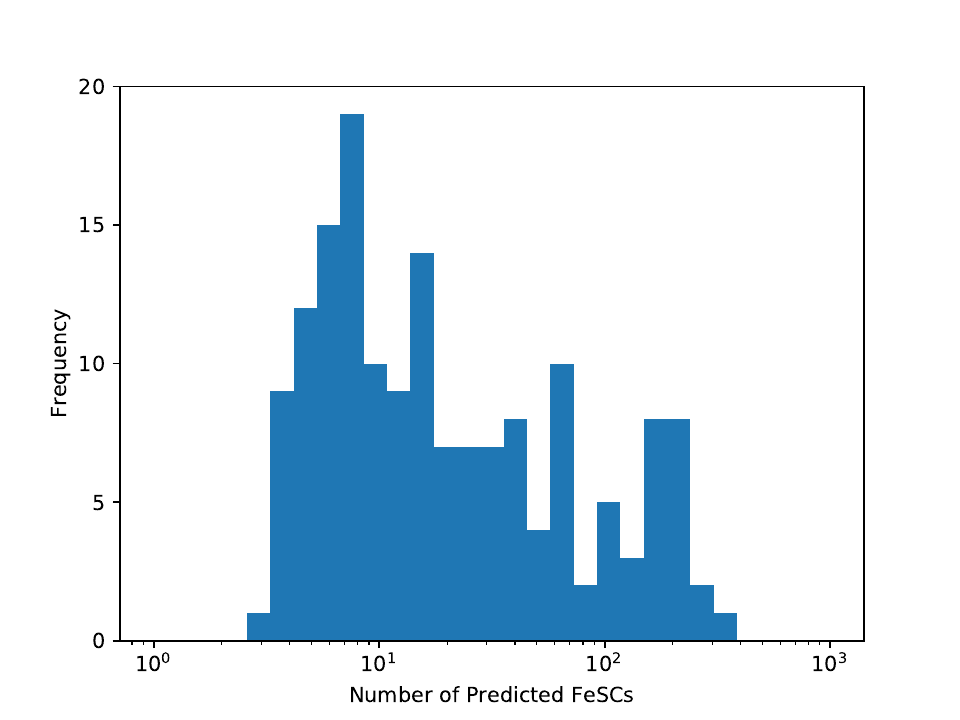}
            \caption{Trained with conventional superconductors.}\label{fig: fe-based-2018-a}
    \end{subfigure}%
    \begin{subfigure}[t]{0.5\textwidth}
        \centering
        \includegraphics[width=1.1\textwidth]{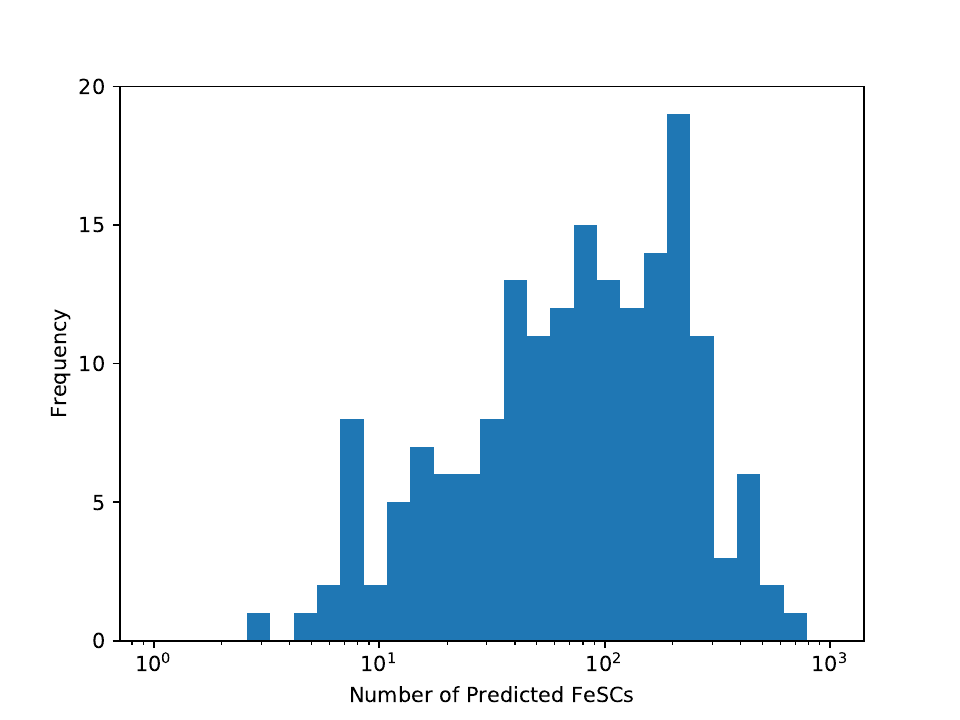}
            \caption{Trained with conventional and cuprate superconductors.}\label{fig: fe-based-2018-b}
    \end{subfigure}
    \caption{Histograms of predicted FeSCs (log scale).}
    \label{fig: fe-based-2018}
\end{figure}

These results show that these two materials had a large impact on the predictions. It is surprising, in view of deep learning, that 2 out of 60,000 training data points had such a significant influence on the model. However, this influence is reasonable because human experts can infer many FeSCs if they know that LaFePFO and LaFePO are superconductors.
The results for (iii) and (v) are shown in Fig.~\ref{fig: fe-based-2018}. Model (iv) (trained with only cuprates) could not predict any FeSCs. As shown in the results of (iii) and (v), the average number of the FeSCs predicted to have finite $T_\text{c}$ increased from 46 to 123 when cuprates were included in the training data. The median increased from 15 to 80. %
The model learned the feature of cuprates, and the number of predicted FeSCs was increased by the addition of cuprates.
We checked the models using the materials reported by Hosono et al. We removed materials containing Fe from the materials on the list. The baseline random precision was 0.2. The mean precision values were 0.41 and 0.35, respectively, for models (iii) and (v), indicating that the models were valid since they are three times higher than the baseline value. %

The above results show that the addition of training data increased the number of FeSCs with finite $T_\text{c}$. We conclude that the model learned the feature of superconductivity. This confirms the reliability of the model, which was also confirmed by checking the precision of the prediction of superconductivities from the materials reported by Hosono et al.%

\subsection{Prediction of superconductivity in cuprates} 

\begin{figure}[htpb]
    \begin{subfigure}[t]{0.5\textwidth}
        \centering
        \includegraphics[width=1.1\textwidth]{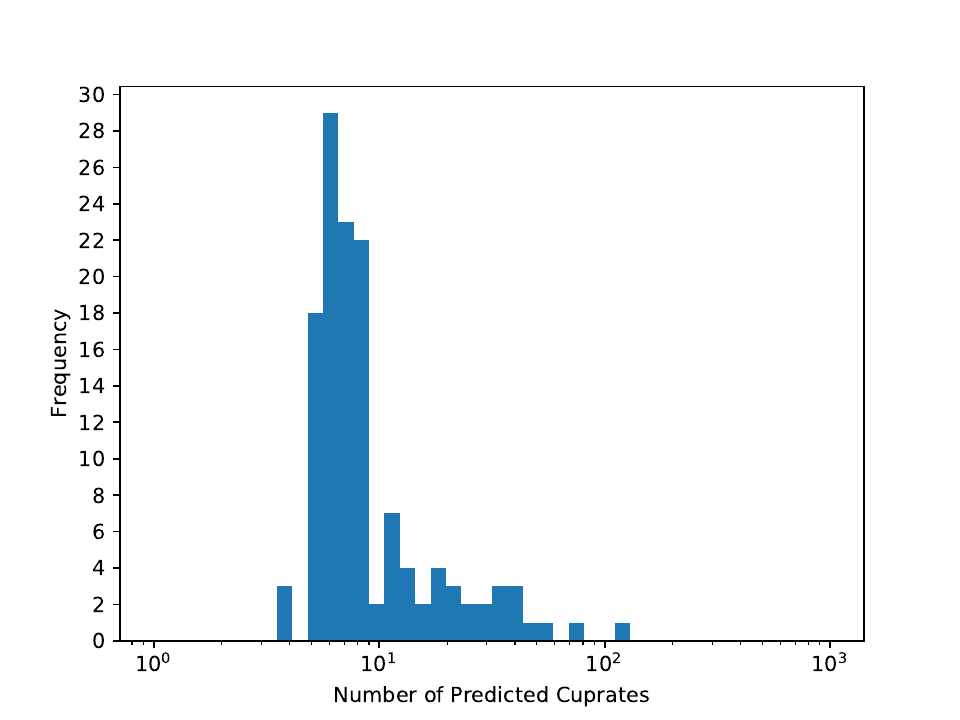}
            \caption{Trained with conventional superconductors.}\label{fig: cuo-based-2018-a}
    \end{subfigure}%
    \begin{subfigure}[t]{0.5\textwidth}
        \centering
        \includegraphics[width=1.1\textwidth]{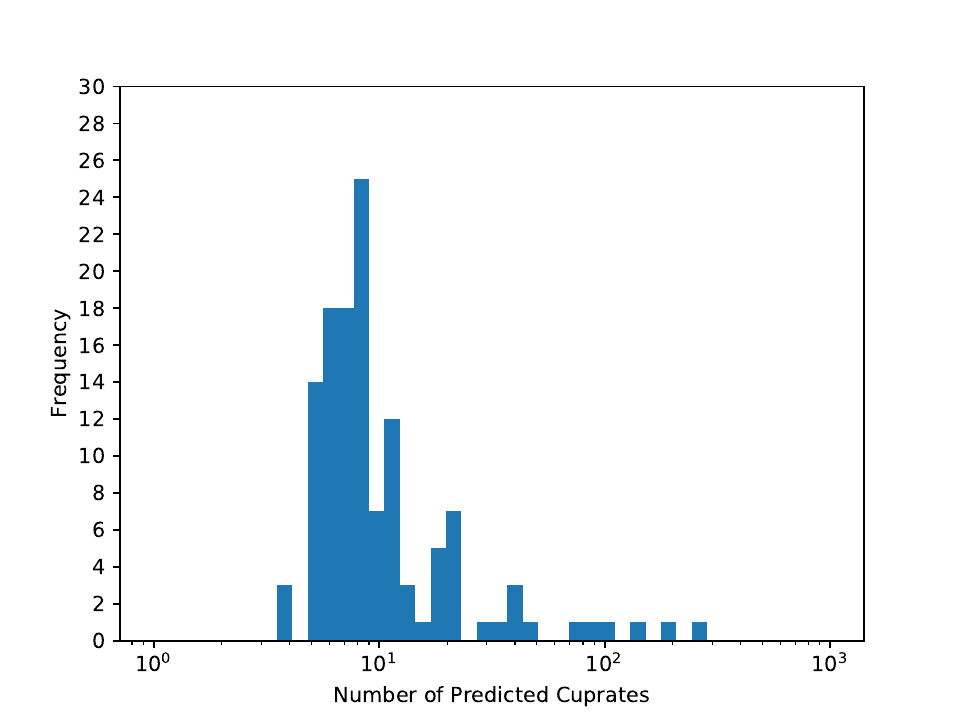}
            \caption{Trained with conventional and Fe-based superconductors.}\label{fig: fe-based-2018-b}
    \end{subfigure}
    \caption{Histograms of predicted cuprate superconductors (log scale).}
    \label{fig: cuo-based-2018}
\end{figure}

It was checked whether the deep learning model could predict cuprates when trained with data that did not include cuprates. We trained models with the following data combinations: (i) only conventional superconductors; (ii) only FeSCs; and (iii) both conventional superconductors and FeSCs. The data were as of 2018. We repeated the training and test approximately 130 times.
 The results for models (i) and (iii) are shown in Fig.~ \ref{fig: cuo-based-2018}. Model (ii) (trained with only FeSCs) could not predict any cuprates with finite $T_\text{c}$, which is consistent with the prediction of FeSCs from cuprates. The average numbers of cuprates predicted to have finite $T_\text{c}$ were 12 and 17 for models (i) and  (iii), respectively. The median values were 7 and 9, respectively. In contrast to the prediction of FeSCs, the mean and median values were almost unchanged by the addition of cuprates to the training data. In SuperCon, the number of FeSCs is less than a quarter of the number of cuprates, which might have led to the difference between the predictions of FeSCs and cuprates. The models were checked using the materials reported by Hosono et al. The baseline random precision was 0.20. The mean precision values were 0.41 and 0.52 for models (i) and (iii), respectively. The precision values are sufficiently higher than the baseline random precision. Although care should be taken when interpreting the results (i.e., deep learning predicted cuprates as candidate superconductors), the results show the possibility of exploring superconductors using deep learning.

\section{Deep learning regression model and binary classification model}
The difference between the deep learning regression model and binary classification model is that the regression model outputs $T_\text{C}$, whereas the binary classification model classifies materials with respect to whether the $T_\text{C}$ value is larger than a threshold value or not.

\section{The features used in random forest classification}
The 32 basic features of elements used for random forest classification are as follows. 
\begin{quotation}
AtomicWeight, Column, DipolePolarizability, FirstIonizationEnergy, GSbandgap,
GSenergy-pa, GSestBCClatcnt, GSestFCClatcnt, GSmagmom, GSvolume-pa, 
ICSDVolume, IsAlkali, IsDBlock, IsFBlock, IsMetal, IsMetalloid, IsNonmetal, MendeleevNumber, 
NdUnfilled, NdValence, NfUnfilled, NfValence, NpUnfilled, NpValence, NsUnfilled, 
NsValence, Number, NUnfilled, NValance, Polarizability, Row, FirstIonizationEnergies.
\end{quotation}
We used the weighted average, weighted variance, maximum, minimum, range, mode, median, and mean absolute difference of the basic features. Thus, in total, we used 256 features.\\

\end{document}